\documentclass[11pt]{article}

\usepackage[preprint]{acl}
\usepackage{times}
\usepackage{latexsym}
\usepackage[T1]{fontenc}
\usepackage[utf8]{inputenc}
\usepackage{microtype}
\usepackage{inconsolata}
\usepackage{graphicx}
\usepackage{multirow}
\usepackage{tcolorbox}
\tcbuselibrary{skins}
\usepackage{pifont}
\usepackage{subcaption}
\usepackage{soul}
\usepackage{nicematrix}
\usepackage{colortbl}
\newcommand{\legend}[1]{\vspace{-0.25cm} \caption*{\scriptsize {#1}} \vspace{-0.25cm}}
\newcommand{\cmark}{\ding{51}}
\newcommand{\xmark}{\ding{55}}
\newtcolorbox{cornerbox}[2]{
  enhanced,
  attach boxed title to top left={
    xshift=12pt,
    yshift=-6pt, 
  },
  colback=#1,
  colframe=black,
  fonttitle=\small\sffamily,
  coltitle=black,
  boxed title style={
    colback=white,
    colframe=black,
    rounded corners,
  },
  top=8pt, 
  title=#2,
}

\title{Evaluation of Clinical Trials Reporting Quality \\ using Large Language Models \\ \small (English translation)}

\author{
 \textbf{Mathieu Laï-king\textsuperscript{*}},
 \textbf{Patrick Paroubek\textsuperscript{*}},
\\
 \textsuperscript{*}Université Paris-Saclay, CNRS, LISN, 91400 Orsay,
\\
 \small{
   \textbf{Correspondence:} \href{mailto:laiking.mathieu@gmail.com}{laiking.mathieu@gmail.com} , \href{mailto:patrick.paroubek@lisn.upsaclay.fr}{patrick.paroubek@lisn.upsaclay.fr}
 }
}

\begin{document}
\maketitle
\begin{abstract}
Reporting quality is an important topic in clinical trial research articles, as it can impact clinical decisions. In this article, we test the ability of large language models to assess the reporting quality of this type of article using the Consolidated Standards of Reporting Trials (CONSORT). We create \textsc{CONSORT-QA}, an evaluation corpus from two studies on abstract reporting quality with CONSORT-abstract standards\footnote{we publicly release our corpus here : \url{https://huggingface.co/datasets/laiking/consort-qa}}. We then evaluate the ability of different large generative language models (from the general domain or adapted to the biomedical domain) to correctly assess CONSORT criteria with different known prompting methods, including \textit{Chain-of-thought}. Our best combination of model and prompting method achieves 85~\% accuracy. Using \textit{Chain-of-thought} adds valuable information on the model's reasoning for completing the task\footnote{The code for our experiments is available at : https://github.com/mathieulaiking/consort-qa}.
\end{abstract}

\begin{figure*}[ht!]
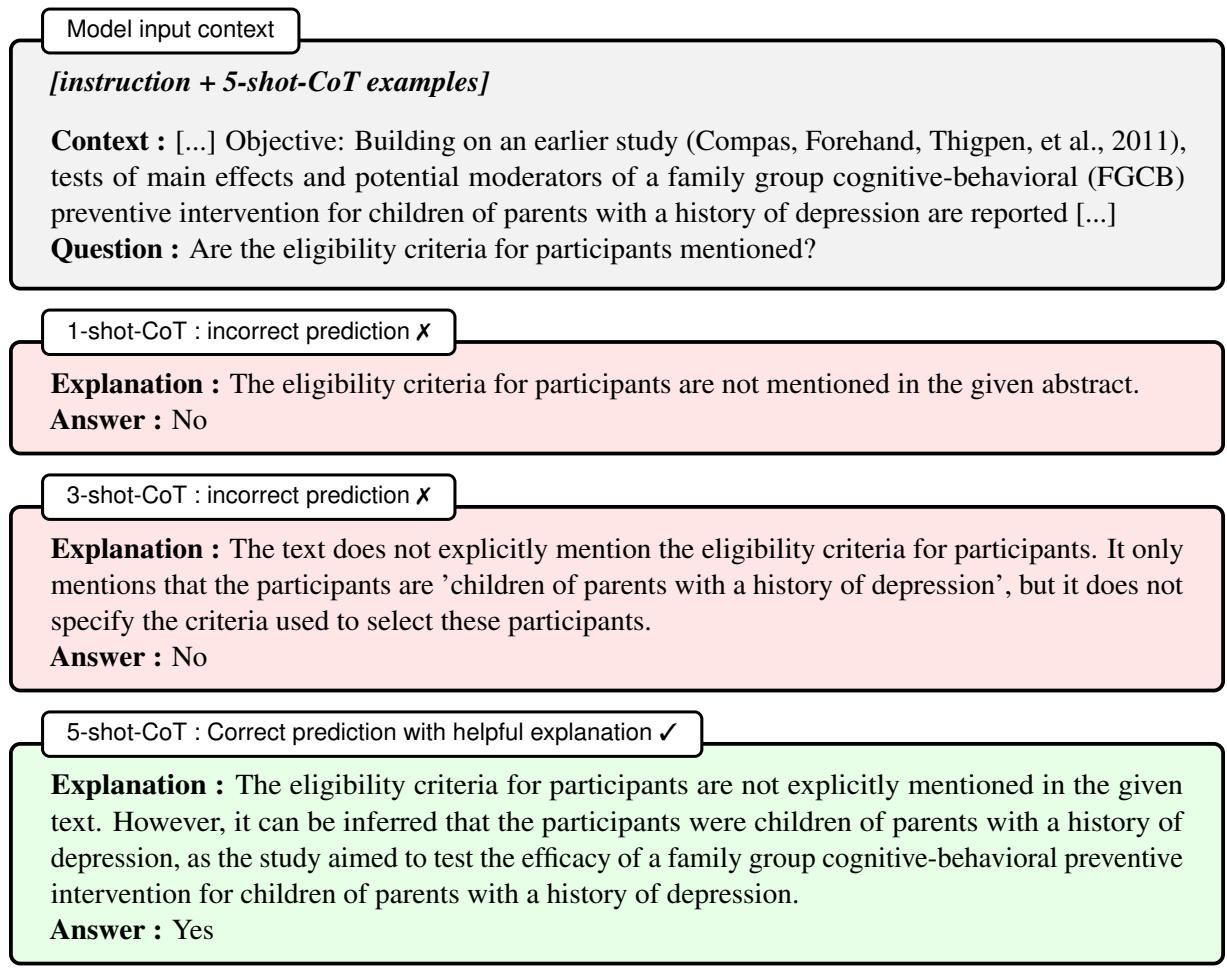

    \centering
    \begin{cornerbox}{gray!10}{Model input context}
    \textit{\textbf{[instruction + 5-shot-CoT examples]}} 
    \vspace{0.3cm}
    
    \textbf{Context :} [...] Objective: Building on an earlier study (Compas, Forehand, Thigpen, et al., 2011), tests of main effects and potential moderators of a family group cognitive-behavioral (FGCB) preventive intervention for children of parents with a history of depression are reported [...]
    
    \textbf{Question :} Are the eligibility criteria for participants mentioned?
    \end{cornerbox}
    
    \begin{cornerbox}{red!10}{1-shot-CoT : incorrect prediction \xmark}
    \textbf{Explanation :} The eligibility criteria for participants are not mentioned in the given abstract.
    
    \textbf{Answer :} No 
    \end{cornerbox}
    
    \begin{cornerbox}{red!10}{3-shot-CoT : incorrect prediction \xmark}
    \textbf{Explanation :} The text does not explicitly mention the eligibility criteria for participants. It only mentions that the participants are 'children of parents with a history of depression', but it does not specify the criteria used to select these participants.
    
    \textbf{Answer :} No 
    \end{cornerbox}
    
    \begin{cornerbox}{green!10}{5-shot-CoT : Correct prediction with helpful explanation \cmark}
    \textbf{Explanation :} The eligibility criteria for participants are not explicitly mentioned in the given text. However, it can be inferred that the participants were children of parents with a history of depression, as the study aimed to test the efficacy of a family group cognitive-behavioral preventive intervention for children of parents with a history of depression.
    
    \textbf{Answer :} Yes 
    \end{cornerbox}

    \caption{Example of generations for our "Few-Shot Chain-of-Thought" strategy with 1,3, or 5 examples using the Mixtral-8x22B-Instruct model for an abstract of the CONSORT-QA-Depression corpus annotated by experts as correctly reported for the eligibility mention criterion}
    \label{fig:cot135-example-4}
\end{figure*}

\section{Introduction}
{\em This article is an English translation of an article published in French in the journal\footnote{\url{https://www.atala.org/revuetal}} "Traitement  Automatique des Langues (TAL)" in 2024 \citep{laiking-paroubek-consort-qa}. Some parts have been modified compared to the original article to improve clarity and transparency.}\\

Clinical trials, more particularly Randomized Controlled Trials (RCTs), are considered the gold standard for assessing the efficacy of a treatment. In general, scientists report RCTs in scientific publications after the end of the study. For the medical community to be able to take appropriate and informed decisions, the articles reporting an RCT must mention all characteristics (methods and results) of the study in a clear and precise way.

Standards that specify which elements to report and how to do it exist to ensure the quality of RCT description in articles. The \textit{Consolidated Standards of Reporting Trials} (\mbox{CONSORT}) are currently the most widely used standards. They have been established in 1996 \citep{beggImprovingQualityReporting1996} and have been updated twice, first in 2001 \citep{altmanRevisedCONSORTStatement2001}, then in 2010 \citep{moherCONSORT2010Explanation2010}.
In what follows, we refer to the last version with the term \mbox{\textit{CONSORT-2010}}. Note that various extensions of the standards have been developed depending on the clinical trial type or for specific sections of an article, e.g., for the abstract \citep{hopewellCONSORTReportingRandomized2008}. 

Although the quality of scientific articles has improved with time since the publication of the norms, some studies reported that it is still insufficient in several medical domains \citep{turnerDoesUseCONSORT2012,warrierCompletenessReportingOutcome2022,wangAbstractsReportsRandomized2021}.

From the perspective of natural language processing, the arrival of \textit{Transformers} \citep{vaswaniAttentionAllYou2017a} has opened up new application possibilities. In particular, using pre-trained generative models based on \textit{Transformers} has grown rapidly in recent years. These models can solve many text analysis and generation tasks \citep{brownLanguageModelsAre2020a,touvronLLaMAOpenEfficient2023,touvronLlamaOpenFoundation2023}. Furthermore, fine-tuning this type of model for the biomedical domain shows new performance on domain-specific tasks \citep {luoBioGPTGenerativePretrained2022,singhalLargeLanguageModels2023}, such as the MedQA question-answering corpus \citep{jinWhatDiseaseDoes2021a}, which includes medical questions similar to those found on the USMLE exam\footnote{United States Medical Licensing Examination.}. Today, it is even possible to achieve expert-level performance in question-answering tasks in the biomedical field with extensive models such as Med-PalM 2 \citep{singhalExpertLevelMedicalQuestion2025}.

That is why we are studying the ability of large generative language models to automatically produce a CONSORT assessment of articles reporting a clinical trial, focusing in this study on the CONSORT checklist for article abstracts. Our contributions include:
\begin{itemize}
\item adapting two English corpora from CONSORT assessments carried out by expert annotators for RCTs concerning COVID-19 interventions on the one hand, and depression prevention in children and adolescents on the other, to the question-answering task;
\item the evaluation of different variants of large public generative language models on this task using different in-context learning methods (also known as prompting methods). The study relies on generative models without specific training for the task due to the small size of the corpora, which we only use for evaluation.
\end{itemize}
We highlight an example of our best model with the best in-context learning method in figure~\ref{fig:cot135-example-4}.

\section{Related work}

\paragraph{Information extraction in clinical trials: PICO, Risk of bias} In the past, researchers in NLP have addressed several tasks concerning articles reporting clinical trials. First, we note information extraction tasks, for instance, with the extraction of PICO entities (\textit{Population, Intervention, Comparator, Outcome}), which are the key elements to detect in a clinical trial report \citep{jinPICOElementDetection2018,mutindaAutomaticDataExtraction2022,wangPICOEntityExtraction2022}. There were also some work done on the automatic evaluation of risk of bias in articles reporting RCTs. Human annotators verify these articles with the help of guidelines like the ones from Cochrane: \textit{Risk of Bias 2} \citep{sterneRoBRevisedTool2019}. \citet{marshallAutomatingRiskBias2015}  and  \citet{marshallRobotReviewerEvaluationSystem2016} have developed a complete system relying on machine learning for assessing risk of bias in an article. More recently, methods based on deep neural transformer architecture, and more specifically  BERT \citep{devlinBERTPretrainingDeep2019}, have been applied to pre-clinical literature for evaluating the risk of bias \citep{wangDevelopmentValidationNatural2020}. However, people who do this type of analysis are still reluctant to adopt this kind of automatic approach in a standard evaluation procedure \citep{jardimAutomatingRiskBias2022}.

\paragraph{Spin detection} We also want to mention works on the detection of "spin", which is the inadequate presentation of research results \citep{IBoutron2014}.  \citep{korolevaAssistedAuthoringAvoiding2020} have investigated different information extraction approaches based on transformers to identify the relation between the presentation of an RCT result and its statistical significance.
See also more recent works on the same issue, since spin is often an object of study in research on clinical trial reporting quality \citep{beroCrosssectionalStudyPreprints2021,wangAbstractsReportsRandomized2021}. 

\paragraph{CONSORT standard automatic assessment} We note several studies that have examined aspects of automatic information extraction for CONSORT standards. \citet{kilicogluAssessingClinicalTrial2021} created the CONSORT-TM corpus, consisting of sentences from articles annotated with CONSORT-2010 elements, and attempted to predict whether a sentence helps determine a CONSORT criterion using multi-label classification modeling. To do this, they use BERT-type models \citep{devlinBERTPretrainingDeep2019} adjusted to their corpus. They also extend this work and test methods using generative models such as the OpenAI GPT-4 model \citep{jiangCONSORTTMTextClassification2024} with in-context learning, or the BioGPT model \citep{luoBioGPTGenerativePretrained2022} trained on their corpus. However, these new approaches do not provide improvements compared to PubMedBERT models \citep{}. Furthermore, this corpus was annotated solely for extracting sentences related to a criterion, not for evaluating whether the criterion is verified, which, in our case, can be important. Moreover, using the \textit{Chain-of-Thought} mechanism adds explainability to the answer the model provides, which is a desirable property in the context of a human annotator's assistant, for example.

\paragraph{In-context learning or prompting} With recent progress of generative LLMs, research on prompting methods have produced surprising results, for instance the \citet{kojimaLargeLanguageModels2022a} study showed that adding a simple sentence like "\textit{Let's think step by step}" to a prompt for eliciting a chain of thought reasoning (COT) of the model resulted in a considerable improvement of the model capacity to perform logical reasoning in various tasks. Recent works have improved this method by optimizing the prompts expressed directly in natural language \citep{yangLargeLanguageModels2023}. Finally, \citet{weiChainofThoughtPromptingElicits2022} has successfully experimented with adding an example of reasoning from the prompt to the answer before stating that the effective prompt can yield better performance for several generative models.
This method is also interesting for question answering tasks in the biomedical domain \citep{singhalExpertLevelMedicalQuestion2025}.

\section{Corpus}\label{sec:corpus}

\begin{table*}[!ht]
    \begin{center}
        \begin{tabular}{|l|p{1.3cm}|p{1.5cm}|p{1.7cm}|p{2cm}|}
            \hline
            \textbf{sujet du corpus} & \textbf{nombre total de résumés} & \textbf{mots par résumé (moy.)} & \textbf{sections par résumé (moy.)} & \textbf{ratio total de réponses oui/non} \\
            \hline
            Dépression & 99 & 225.90 & 3.69 & 0.30/0.70 \\\hline
            COVID-19 & 40 & 339.50 & 5.58 & 0.48/0.52 \\
            \hline
        \end{tabular}
        \caption{General statistics for the corpora used (avg. = average). Averages are calculated based on all abstracts in the corpus.}
        \label{tab:corpora_stats}
    \end{center}
\end{table*}

We have created an annotated corpus based on existing high-quality studies on CONSORT standards (i.e., with annotations made by experts). However, many full-text articles are not publicly available in clinical research. We therefore chose to limit our research to CONSORT standards for abstracts \citep{hopewellCONSORTReportingRandomized2008}, which we will refer to as \textsc{CONSORT-abstract} in the rest of this chapter. This has three advantages:
\begin{itemize}
    \item abstracts must still comply with standards, as access is also limited for some doctors who do not necessarily have a subscription to all scientific journals publishing clinical trial reports, or simply in cases where some experts are content to read the abstract of a clinical trial report.
    \item Access to abstracts is also easier for creating our corpus (because we need to retrieve the text of each article in the studies considered, and we do not have unlimited access to retrieve full texts of articles that may come from different paid scientific journals). 
    \item For our purpose of using neural language models and in-context learning strategies, it is necessary to present examples that do not exceed the maximum input size of these models. Thus, using abstracts rather than the full text allows this limitation to be circumvented. 
\end{itemize}

\subsection{Corpus adaptation}\label{sec:corpus-adaptation}
First, we aimed to have a corpus with several abstracts written by experts and associated with their human evaluation according to \mbox{CONSORT-abstract} criteria. We searched the public archive PubMedCentral\footnote{\url{https://www.ncbi.nlm.nih.gov/pmc/}} for articles mentioning the observance of \mbox{CONSORT-abstract} criteria. Our search query was "adherence to \mbox{CONSORT-abstract}". Then, we have reviewed each study, including the supplementary material section and the data availability statement, to determine whether the authors released the detailed evaluation results for each abstract presented in their research. We limited our manual review to the first forty articles of the list returned by the PubMedCentral search engine. Among the forty articles reviewed, only two (5\%) validated our requirement about the availability of the study's detailed results. Our corpus comprises the evaluation quality data from these two remaining studies.

Since the authors of this kind of study did not intend their data to be the subject of an automatic analysis, they provide in general only a DOI\footnote{DOI (\textit{Digital Object Identifier}) is a normalized numeric identifier of an object accessible in Internet (physical, electronic or abstract):\url{https://www.doi.org/}} or a title and some metadata for the abstracts that they annotated, but not its content (text). Consequently, we had to search for the text of the abstracts, using their PubMed identifiers when they were available and otherwise using the \mbox{PubMed IdConverter} API\footnote{\url{https://www.ncbi.nlm.nih.gov/pmc/tools/idconv/}} to map the DOI onto the corresponding PubMed identifier and collect the abstract via the Entrez API.

\subsection{Corpus description}\label{sec:corpus-description}

\begin{table*}[ht!]
    \resizebox{\textwidth}{!}{%
    \begin{tabular}{|c|c|cc|}
    \hline
    \textbf{d\_id} & \textbf{c\_id} & \multicolumn{1}{c|}{\textbf{d\_question}} & \textbf{c\_question} \\ \hline
    D01 & C01 & \multicolumn{2}{c|}{Is the study identified as randomized in the title ?} \\ \hline
    D02 & C02 & \multicolumn{2}{c|}{Is there a structured summary of the trial design (e.g., parallel, crossover, cluster, non-inferiority) ?} \\ \hline
    D03a & C03a & \multicolumn{2}{c|}{Are the eligibility criteria for participants mentioned?} \\ \hline
    D03b & C03b & \multicolumn{2}{c|}{Are the settings or locations where the data were collected stated in the abstract ?} \\ \hline
    D04a & \multirow{2}{*}{C04} & \multicolumn{1}{c|}{\begin{tabular}[c]{@{}c@{}}Do the authors report essential features of the \\ experimental intervention (if needed) ?\end{tabular}} & \multirow{2}{*}{\begin{tabular}[c]{@{}c@{}}Are the interventions sufficiently \\ detailed for each group (eg, when, how) ?\end{tabular}} \\ \cline{1-1} \cline{3-3}
    D04b &  & \multicolumn{1}{c|}{\begin{tabular}[c]{@{}c@{}}Do the authors report essential features of the\\  comparison (= control) intervention (if needed) ?\end{tabular}} &  \\ \hline
    D05 & C05 & \multicolumn{1}{c|}{Are there specific objectives or hypothesis stated ?} & Are there specific objectives or hypothesis stated ? \\ \hline
    D06a & \multirow{2}{*}{C06} & \multicolumn{1}{c|}{\begin{tabular}[c]{@{}c@{}}Do the authors explicitly state the primary \\ outcome as such (eg, primary / main / principal) ?\end{tabular}} & \multirow{2}{*}{\begin{tabular}[c]{@{}c@{}}Are the primary outcomes clearly \\ described for this trial in methods ?\end{tabular}} \\ \cline{1-1} \cline{3-3}
    D06b &  & \multicolumn{1}{c|}{\begin{tabular}[c]{@{}c@{}}Do the authors explicitly state when\\  the primary outcome was assessed (time frame) ?\end{tabular}} &  \\ \hline
     & C07a & \multicolumn{1}{c|}{} & \begin{tabular}[c]{@{}c@{}}Is the random assignment declared (eg, random, \\ randomized, randomization, random allocation) ?\end{tabular} \\ \hline
    D07 & C07b & \multicolumn{2}{c|}{\begin{tabular}[c]{@{}c@{}}If they declared a random allocation to interventions (if they did not answer no), do the authors \\ correctly report how they were allocated (e.g., computer-generated, random numbers, coin toss, etc.) ?\end{tabular}} \\ \hline
     & C07c & \multicolumn{1}{c|}{} & Are they referring to allocation concealment ? \\ \hline
    D08a & \multirow{3}{*}{C08} & \multicolumn{1}{c|}{\begin{tabular}[c]{@{}c@{}}Do authors describe if participants were blinded ?\\ (answer yes only if participants are blinded, \\ do not care about caregivers or outcome assessors)\end{tabular}} & \multirow{3}{*}{\begin{tabular}[c]{@{}c@{}}Are they mentioning whether or not participants,\\ trial providers and data collectors were blinded ?\end{tabular}} \\ \cline{1-1} \cline{3-3}
    D08b &  & \multicolumn{1}{c|}{\begin{tabular}[c]{@{}c@{}}Do authors describe if the program deliverer\\  (caregiver) were blinded ? (answer yes only\\  if caregivers are blinded, do not care \\ about participants or outcome assessors)\end{tabular}} &  \\ \cline{1-1} \cline{3-3}
    D08c &  & \multicolumn{1}{c|}{\begin{tabular}[c]{@{}c@{}}Do authors describe if data collectors (outcome\\  assessors, analysts) were blinded ? (answer yes \\ only if data collectors are blinded, do \\ not care about participants or caregivers)\end{tabular}} &  \\ \hline
     & C08a & \multicolumn{1}{c|}{} & \begin{tabular}[c]{@{}c@{}}Is there only a brief description of blinding \\ (eg, single-blind, double-blind, triple-blind) ?\end{tabular} \\ \hline
    D09 & C09 & \multicolumn{2}{c|}{Are the numbers of participants randomized to each group clearly stated ?} \\ \hline
     & C10 & \multicolumn{1}{c|}{} & \begin{tabular}[c]{@{}c@{}}Is the Trial status (eg, on-going, closed to recruitment, \\ closed to follow-up, etc.) mentionned ?\end{tabular} \\ \hline
    D10 & C11 & \multicolumn{2}{c|}{\begin{tabular}[c]{@{}c@{}}Are the numbers of participants analyzed for each group clearly stated \\ (not the number randomized but the patients included in the analysis of the primary outcome) ?\end{tabular}} \\ \hline
     & C11a & \multicolumn{1}{c|}{} & \begin{tabular}[c]{@{}c@{}}Are the numbers of participants analyzed in \\ accordance with the original grouping (eg, \\ intention-to-treat analysis or pre-protocol analysis) ?\end{tabular} \\ \hline
    \multirow{3}{*}{D11} & C12a & \multicolumn{1}{c|}{\multirow{3}{*}{\begin{tabular}[c]{@{}c@{}}For the primary outcome, \\ is there a result for each group and\\  the estimated effect size and its\\  precision (e.g., 95\% CI ) ? \\ (if one of them is missing, answer no)\end{tabular}}} & \begin{tabular}[c]{@{}c@{}}For the primary outcome(s) ,\\ is there a summary report of results for each group ?\end{tabular} \\ \cline{2-2} \cline{4-4} 
     & C12b & \multicolumn{1}{c|}{} & \begin{tabular}[c]{@{}c@{}}For the primary outcome(s), \\ is the estimated effect size clearly stated ?\end{tabular} \\ \cline{2-2} \cline{4-4} 
     & C12c & \multicolumn{1}{c|}{} & \begin{tabular}[c]{@{}c@{}}For the primary outcome(s), is the precision \\ of the estimate (eg, 95\%CI) clearly stated ?\end{tabular} \\ \hline
    D12 & C13 & \multicolumn{2}{c|}{Do the authors correctly mention the presence or absence of adverse events or side effects ?} \\ \hline
    D13a &  & \multicolumn{1}{c|}{Do the authors state the conclusions of the trial?} &  \\ \hline
    D13b &  & \multicolumn{1}{c|}{\begin{tabular}[c]{@{}c@{}}Do the authors state implications \\ for further research or clinical practice ?\end{tabular}} &  \\ \hline
     & C14 & \multicolumn{1}{c|}{} & Are the general interpretations corresponding to the results ? \\ \hline
     & C14a & \multicolumn{1}{c|}{} & Are the benefits and harms balanced in the conclusion ? \\ \hline
    D14a & \multirow{2}{*}{C15} & \multicolumn{1}{c|}{Do the authors provide the trial registration number ?} & \multirow{2}{*}{\begin{tabular}[c]{@{}c@{}}Are the trial registration number and the name of trial \\ register clearly stated ? Answer no if one of them is missing\end{tabular}} \\ \cline{1-1} \cline{3-3}
    D14b &  & \multicolumn{1}{c|}{Do the authors provide the name of the trial register ?} &  \\ \hline
    D15 & C16 & \multicolumn{1}{c|}{Do the authors declare the source of funding ?} & Do the authors declare the source of funding ? \\ \hline
    \end{tabular}%
    }
    \caption{Correspondence between CONSORT questions/criteria for the \textnormal{\textsc{consort-qa-depression}} corpus (prefix “d”) and for the \textnormal{\textsc{consort-qa-covid}} corpus (prefix “c”)}
    \label{tab:consort-questions}
\end{table*}

Our corpus \textsc{consort-qa}  is composed of two sub corpora, one for each of the two  articles that resulted from our search of PubMed:
\begin{itemize}
    \item \textbf{\textsc{consort-qa-covid}}:  \citet{wangAbstractsReportsRandomized2021} have published a study about the observance of \mbox{CONSORT-abstract} in articles about \mbox{COVID-19} RCTs. With the \textit{Entrez} API, we collected forty (40) abstracts evaluated according to sixteen (16) criteria of the \mbox{CONSORT-abstract} list with a boolean measure (validated or not).
    \item \textbf{\textsc{consort-qa-depression}}: \citet{wiehnReportingQualityAbstracts2022} presented a study concerning the prevention of children or adolescent depression. In addition to RCTs, the authors considered \textit{Cluster Randomized trials}. For this study's cluster trials, we could automatically retrieve the sixty (63) abstracts evaluated. For the RCTs, we collected with the \textit{Entrez} API eighty-four (84) abstracts automatically over the hundred and three (103) abstracts considered by this study. Then, with a Google search using either DOIs or titles, we added fifteen (15) missing abstracts for a total of ninety-nine (99) RCT abstracts retrieved and four (4) still missing. Overall, we got one hundred and sixty-two (162) abstracts, cluster trials and RCTs. In this depression study, the authors used a different annotation scheme for the \mbox{CONSORT} criteria evaluation, which required some harmonizing on our part with the one used in the COVID study. Concerning the criterion about funding sources of the trial, the evaluators identified the complete text span referring to the source. They augmented the set of evaluation values with the specific mention "\textit{in another section}" when the criterion was validated but the source mentioned was in the body of the article and not in the abstract. For these particular cases, we decided to consider that the criterion was not validated since we gave only the text of the abstract as input to our generative models in our experiments. Last, instead of a boolean value for the criteria, the authors considered three levels of precision: "\textit{not reported}", "\textit{inadequately reported}", and "\textit{correctly reported}". We felt justified to merge the first two values into the value "\textit{not reported}" to have boolean annotations for the criteria, since in our experiments, these cases correspond to a criterion that is not validated. Finally, note that in this depression study, the authors provide the values of inter-annotator agreement for each criterion, upon which we base our assessment of the criterion's intrinsic difficulty.
\end{itemize}

Concerning the \textsc{CONSORT-abstract} criteria evaluated in both corpora, although these criteria are derived from the same standard, some of them have been slightly adapted to suit the needs of the evaluators, for example, by subdividing specific criteria or even adding a sub-criterion to provide a more transparent and more precise annotation. In addition, the identifier of a particular criterion (numbered with a number and, optionally, a letter: for example, 13a) may also vary depending on the evaluators. To avoid confusion, we add a letter at the beginning of this identifier to differentiate between the two corpora. We present the criteria defined in this way (in the form of questions for the task formulation, see section \ref{sec:task_formulation}) in Table~\ref{tab:consort-questions}. You can note that, compared to the original \textsc{CONSORT-abstract} criteria from \citet{hopewellCONSORTReportingRandomized2008}, some have been divided into several sub-criteria: for example, in \textsc{CONSORT-QA-Depression}, the criterion verifying the secret assignment mechanism is divided into three. This shows that some criteria require several steps of reasoning to arrive at the answer.

We retrieve the inter-annotator agreement values for each criterion from the depression study (which uses a Kappa metric). Using the correspondences between the criteria defined in the \textsc{consort-qa-covid} and \textsc{consort-qa-depression} corpora, we assimilate the agreement values for the criteria defined in the \textsc{consort-qa-covid} study. We use these values to evaluate each criterion's difficulty and determine whether generative language models encounter more difficulty with specific criteria than with others. We present these values in table~\ref{tab:corpus-kappa}.

\begin{table*}[ht]
    \begin{tabular}[t]{|l|l|}
        \hline
        \textbf{id}   & \textbf{kappa} \\ \hline
        D01  & 0.96  \\ \hline
        D02  & 0.38  \\ \hline
        D03a & 0.77  \\ \hline
        D03b & 0.81  \\ \hline
        D04a & 0.80  \\ \hline
        D04b & 0.76  \\ \hline
        D05  & 0.73  \\ \hline
        D06a & 0.91  \\ \hline
        D06b & 0.69  \\ \hline
        D07  & 0.49  \\ \hline
    \end{tabular}
    \quad
    \begin{tabular}[t]{|l|l|}
        \hline
        \textbf{id}   & \textbf{kappa} \\ \hline
        D08a & 0.77  \\ \hline
        D08b & 0.77  \\ \hline
        D08c & 0.66  \\ \hline
        D09  & 0.95  \\ \hline
        D10  & 0.88  \\ \hline
        D11  & 0.94  \\ \hline
        D13a & 0.75  \\ \hline
        D13b & 0.74  \\ \hline
        D14a & 1.00  \\ \hline
        D14b & 0.98  \\ \hline
        D15  & 0.88  \\ \hline
    \end{tabular}
    \quad
    \begin{tabular}[t]{|l|l|}
        \hline
        \textbf{id}   & \textbf{kappa} \\ \hline
        C01  & 0.96  \\ \hline
        C02  & 0.38  \\ \hline
        C03a & 0.77  \\ \hline
        C03b & 0.81  \\ \hline
        C04  & 0.78  \\ \hline
        C05  & 0.73  \\ \hline
        C06  & 0.80  \\ \hline
    \end{tabular}
    \quad
    \begin{tabular}[t]{|l|l|}
        \hline
        \textbf{id}   & \textbf{kappa} \\ \hline
        C07b & 0.49  \\ \hline
        C08  & 0.73  \\ \hline
        C09  & 0.95  \\ \hline
        C12a & 0.94  \\ \hline
        C12b & 0.94  \\ \hline
        C12c & 0.94  \\ \hline
        C15  & 0.99  \\ \hline
        C16  & 0.88  \\ \hline
    \end{tabular}
    \caption{Kappa values defined in the \text{consort-qa-depression} corpus for each criterion. The values for the \text{consort-qa-covid} corpus have been adapted from the values in the other corpus (which is why some identifiers are missing).}
    \label{tab:corpus-kappa}
\end{table*}

The corpora we obtain are relatively small. This is why we decided to conduct our experiments on generative models in inference only, as fine-tuning would have required more data. These corpora are therefore used solely for the evaluation of our methods.

\subsection{Filtering sentences from abstracts in the corpus}\label{sec:sentence_filtering}

We also want to study the influence of the parts of the abstract included in the model's input prompt. We use the work for the CONSORT-TM corpus \citep{kilicogluAssessingClinicalTrial2021} to do this. This corpus consists of fifty full-text articles on randomized controlled trials, which have been annotated at the sentence level to determine whether they contain information on a particular CONSORT criterion (a sentence may be helpful for zero, one, or more criteria).

This corpus has a limitation: it does not contain titles or abstracts and only evaluates CONSORT criteria on methodology. We therefore define a conversion table between the criteria in our corpus and those considered by CONSORT-TM.  

The models we use were trained by the authors of the CONSORT-TM corpus\footnote{Available at the following link: \url{https://github.com/kilicogluh/CONSORT-TM/tree/master/bert}}. These are BioBERT models \citep{lee2020biobert} adjusted to their entire corpus (models used only for inference) on a multi-label classification task. These models exist in two variants: one trained on the sentences of the corpus alone (\textit{text-only-50}) and the other trained on the sentences preceded by the section of text to which they belong (\textit{section-text-50}).  

Knowing that some of the abstracts in our corpora are structured into sections, we first split them according to the identified sections. Then, to separate our abstracts into sentences, we use the Python library Scispacy \citep{neumann-etal-2019-scispacy} with their model \textit{en-core-sci-lg}. This gives us sentences and, optionally, their associated section if available. Finally, to obtain the relevant criteria for each sentence, we use the adjusted models presented above and assign the appropriate criteria to each sentence according to the models' predictions. We use the \textit{text-only-50} model only for sentences that do not have a defined section, and the \textit{section-text-50} model for sentences that belong to a section. 

Suppose the classification does not yield any results for a criterion in an abstract (none of the sentences are associated with the criterion). In that case, the context will consist of the entire abstract, as in the method without filtering.

\section{Methods for automatic verification of CONSORT standard criteria}\label{sec:ch4:methods}

\subsection{Task formulation}\label{sec:task_formulation}

We chose to consider the problem of evaluating an abstract according to the CONSORT criteria as an individual question-answer task for each criterion with a Boolean answer (criterion verified/criterion rejected). To do this, we first manually redefined each criterion in the form of a question. We detail the development of these questions in section~\ref{sec:prompt_development}. Table~\ref{tab:consort-questions} provides the questions for the two corpora. 

We end up with 4,272 abstract/question pairs for our two corpora, which we will provide to the various large generative models selected.

For models adapted to the biomedical field, we use three variants of general models (presented above) adjusted to a biomedical corpus: Meditron \citep{chen2023meditron70b} adjusted from Llama-2, BioMistral \citep{labrakBioMistralCollectionOpenSource2024a} adjusted from Mistral-v0.1, and OpenBioLLM \citep{OpenBioLLMs} adjusted from Llama-3. The data and training processes are not the same for these different models, as there is not necessarily a biomedical variant for all general models, and they are trained independently of each other.

\subsection{In-context learning strategies (or \textup{prompting})}\label{sec:prompting}

\begin{figure*}[!ht]
        \begin{tcolorbox}
            \textbf{Instructions :} The task is to verify a criterion from the Consolidated Standards of Reporting Trial (CONSORT) for a given abstract. The output should be yes or no (whether the criterion is met or not).
        \end{tcolorbox}
        \begin{tcolorbox}
            \textbf{Context :} """Title : A randomized [...]""".\\
            \textbf{Question :} Are the numbers of participants analyzed for each group clearly stated (not the number randomized but the patients included in the analysis of the primary outcome)?\\
            \textbf{Explanation :} In the Results section, the authors report the number of participants analyzed for the primary outcome, saying : '300 were included in the analysis of the primary outcome'. They also precise the numbers for each group '(100 in the acetaminophen group, 100 in the ibuprofen group, and 100 in the codeine group)'. So the numbers of participants analyzed for each group is clearly stated.\\
            \textbf{Answer :} Yes
        \end{tcolorbox}
        \begin{tcolorbox}
            \textbf{Context :} """Title: Online attentional [...]""".\\
            \textbf{Question :} Are the numbers of participants analyzed for each group clearly stated (not the number randomized but the patients included in the analysis of the primary outcome)?\\
            \textbf{Explanation :}
        \end{tcolorbox}
    \legend{Here, article abstracts have been shortened for space reasons, and we provide only one example (but for few-shot strategies, we provide several).}
    \caption{Example prompt for the \textnormal{1-shot-cot-orig} strategy in the \textnormal{\textsc{CONSORT-QA-Depression}} corpus}
    \label{fig:prompt_example}
\end{figure*}

As numerous studies have shown, working on the input of a generative model can significantly improve its performance in solving various tasks \citep{weiChainofThoughtPromptingElicits2022,singhalLargeLanguageModels2023}. Since we have several criteria and questions for each article, we ask the model to answer each question separately. We test the following four strategies.
\begin{itemize}
    \item \textbf{\textit{0-shot}} (or zero-example): this strategy is the simplest and forms the basis for the others. We first give a general instruction passage, which asks, in particular, to answer a question with yes or no, then the context, which is the abstract of the article to be evaluated, and finally the question of the specific criterion to be evaluated. For this method, we limit the generation to a single \textit{token}\footnote{Entity referring to the elements that make up the vocabulary of a language model. A \textit{token} can be a part of a word, a word, a punctuation mark, a symbol, a space, etc.} by choosing the first occurrence of one of the two possible answers (\textit{yes} or \textit{no} since the abstracts and prompts are written in English) in the distribution provided by the model, among the twenty most probable \textit{tokens}. If none of the response tokens are found, we consider that the model is hallucinating (and this will therefore be counted as a wrong answer during evaluation).
    
    \item \textbf{\textit{few-shot}}: for this strategy, we add a few examples from the corpora (\textsc{CONSORT-QA-Covid} and \textsc{CONSORT-QA-Depression}) to the prompts from the previous strategy, including (for each instance) the full abstract of the article, the question being evaluated, and the answer. For this purpose, we exclude five examples per corpus. We test three configurations: \textit{1-shot}, \textit{3-shot}, and \textit{5-shot}, in particular because these are the values commonly used and generally achieve the best performance for question-answering tasks with \textit{prompting} \citep{singhalExpertLevelMedicalQuestion2025}. These examples come from the same corpus and contain the same question as the abstract to be evaluated (they are placed before it in the prompt). The final response token is generated in the same way as \mbox{previously}.

    \item \textbf{\textit{1-shot-cot-orig}} (for \textit{Chain-of-Thought}): this strategy is similar to the previous one, but we add an explanation before the answer to the prompt examples and we also ask the model to generate an explanation before its answer \citep{weiChainofThoughtPromptingElicits2022}, which generally improves the capabilities of generative models on problems requiring multiple reasoning steps. To generate the explanation, we use greedy decoding (no sampling) until the next line break (end of the explanation) or when the length of the explanation exceeds 200 tokens. The model then generates the response token as before. There is only one example per prompt here. We take a single instance because we retrieve those given in the original CONSORT abstract article by Hopewell et al. (2008), which includes a single example per criterion (always positive, where the criterion is therefore verified). We manually adapt the explanations according to the sub-criteria defined differently between our two sources (table~\ref{tab:consort-questions}). 

    \item \textbf{\textit{few-shot-cot}}: this method is the same as the previous one, but with several examples in the prompt. Since our corpora do not provide textual annotations explaining the annotator's answer choice, we automatically annotate the examples with a language model. We choose the 70-billion-parameter \textit{Llama-3} instruction-tuned model for this to generate these explanations. We provide the correct answer to the model before generating the explanation to guide it towards a more coherent explanation. We do this for the five examples per corpus, as defined in the few-shot strategy. We do not annotate more examples, perform validation, or apply example selection methods to reduce the cost of the experiments.
\end{itemize}

We provide an example of a prompt supplied to the model in Figure~\ref{fig:prompt_example}. We performed our experiments on NVIDIA A100 GPUs (the number depending on the memory required for each model). We use the Python vLLM library for generation \citep{kwon2023efficient}.

\subsection{Prompt Development}\label{sec:prompt_development}

The format of our prompts (model inputs) is similar to that used in the Med-PaLM 2 model article \citep{singhalExpertLevelMedicalQuestion2025}. To redefine the criteria as questions, we use the original wording used in each study\footnote{For \textsc{CONSORT-QA-Covid}, this is Table 2 of their article, and for \textsc{CONSORT-QA-Depression}, Table S7 given in their supplementary material.} on the \textsc{CONSORT-abstract} assessment (presented in section~\ref{sec:corpus-description}). 

In addition, some questions were adjusted, either because we felt they were not clear enough to solve the task (even for a human annotator), or because we observed recurring errors in \textit{0-shot} on the examples given in the original \textsc{CONSORT-abstract} statement by \citet{hopewellCONSORTReportingRandomized2008} (which are not found in the test data). In particular, this resulted in giving the generative models more precise instructions than the formulations provided by the study authors, some of which we felt were incomplete (notably because the study authors only provide tables with short descriptions of each criterion, rather than a precise annotation guide for each criterion). We also use the original definitions of the \textsc{CONSORT-abstract} and \textsc{CONSORT-2010} criteria to add these clarifications~\citep{hopewellCONSORTReportingRandomized2008,moherCONSORT2010Explanation2010}.

\subsection{Evaluation of the performance of our methods}\label{sec:evaluation}

To assess the model's ability to evaluate a criterion based on an abstract, we measure accuracy at the criterion and corpus levels. We consider each example, containing an abstract and a criterion, equally: therefore, a micro-average, giving more importance to the \textsc{CONSORT-QA-Depression} corpus, which includes more examples. 

For models that do not have sufficient context size for \textit{few-shot} strategies, we evaluate them only in \textit{0-shot}.

Furthermore, to evaluate the influence of filtering abstract sentences on prompt efficiency (section~\ref{sec:sentence_filtering}), we reduce the space of criteria for evaluation. As mentioned earlier, the models we use for filtering do not include all the criteria (and sub-criteria) of our two corpora. Thus, for each corpus, we only consider the intersection between the set of criteria predicted by the sentence classification models and the set of criteria of the corpus to which the test example belongs. In addition, we only consider the \textit{Llama-2} model to reduce computational costs\footnote{We feel all the more confident in this choice given the slight differences in performance observed in our initial experiments with filtering.}.

\section{Results}\label{sec:ch4:results}

We first compare the overall performance of the models (i.e., the micro-average as defined in section~\ref{sec:evaluation}). We then discuss the detailed performance by criterion for the model with the best overall performance. We then study the correlation between these criterion-level performances and the difficulty of each criterion. Finally, we manually analyze the consistency of correct responses for Chain-of-Thought-type generations (both correct and incorrect responses from the best model).

\subsection{Analysis of the performance of models and in-context learning strategies}

\begin{figure}[htbp]
    \centering
    \includegraphics[width=\linewidth]{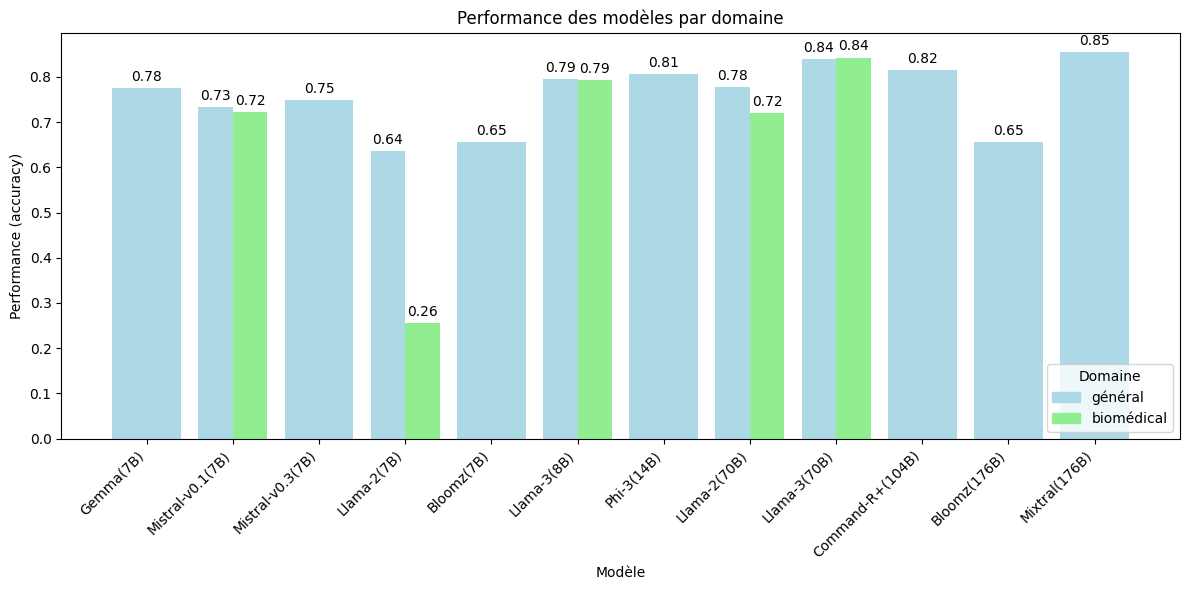}  
    \legend{X-axis : model name and size | Blue bars : General models | Green bars: Biomedical models}
    \caption{Maximum accuracy (best micro-average performance for each in-context learning strategy) achieved for each model tested}
    \label{fig:models_comparison}
\end{figure}

We first compare the different models tested in Figure~\ref{fig:models_comparison}.

\paragraph{Model comparison} Here, as expected, for the same model available in different sizes (such as \textit{Llama-3}, which we evaluate with its eight-billion-parameter version and its 70-billion-parameter version), the larger models perform better. Nevertheless, some smaller models perform very similarly to larger models, or even better for the zero-shot strategy, where Phi-3 achieves the best performance (surpassing models up to 10 times larger in terms of parameters). However, the reason for these differences remains challenging to determine, as the authors of these models do not systematically specify the amount of data used for pre-training and the quality criteria used to select their data. Finally, the model with the best results is the Mixtral-8x22B model adjusted by instructions using the 5-shot-cot method, with an accuracy of 85~\%. These overall performances are encouraging for this type of approach, but show that this task remains difficult, even for the best publicly available models.

\paragraph{Effect of continual pretraining in the biomedical domain} Overall, we observe that the biomedical equivalent of general models (BioMistral, Meditron, and OpenBioLLM) performs less than its general counterpart. This may be due to the nature of the task, which does not necessarily require specific biomedical knowledge (except for certain criteria). In fact, most criteria can be verified simply by searching for information in the provided context, a task for which general language models already perform well. Nevertheless, this difference remains surprising because these biomedical models are trained on data very similar to those provided as input (i.e., biomedical research articles).

\begin{figure}[htbp]
    \centering
    \includegraphics[width=\linewidth]{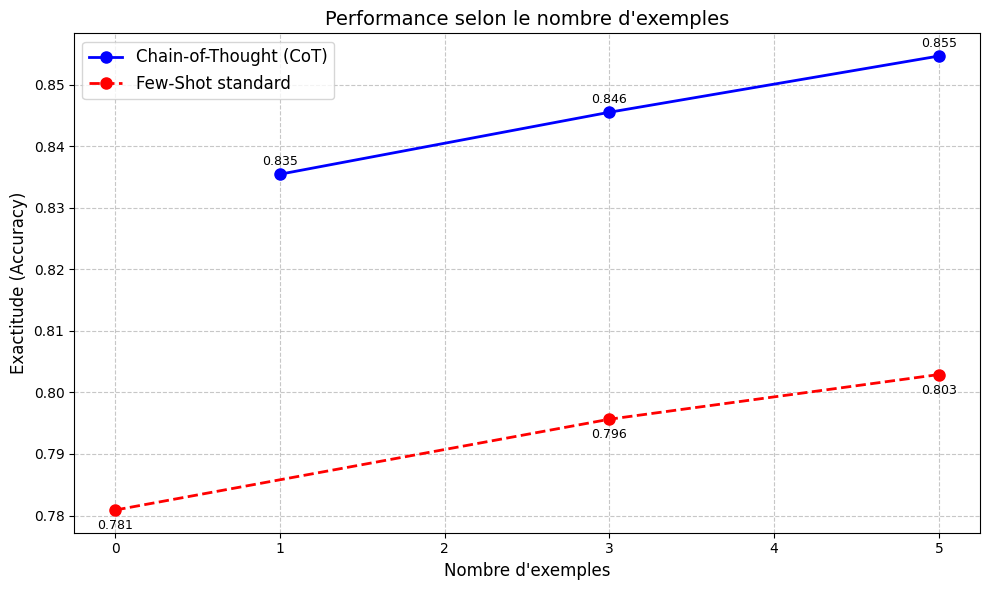}  
    \legend{X-axis: Number of examples}
    \caption{Average accuracy (micro-average on concatenated examples from both corpora) of the Mixtral-8x22B model as a function of the number of examples and according to strategies that add an explanation (CoT) or not}
    \label{fig:few-shot_comparison}
\end{figure}

We then compare the effect of the in-context learning strategies used in Figure~\ref{fig:few-shot_comparison}.

\paragraph{Effect of different in-context learning strategies} Unsurprisingly, we see here that performance increases as more examples are provided in context. Furthermore, adding explanations further improves the model's performance (although it should be noted that the method used to generate the explanation requires more computing time).

\begin{figure}[htbp]
    \centering
    \includegraphics[width=\linewidth]{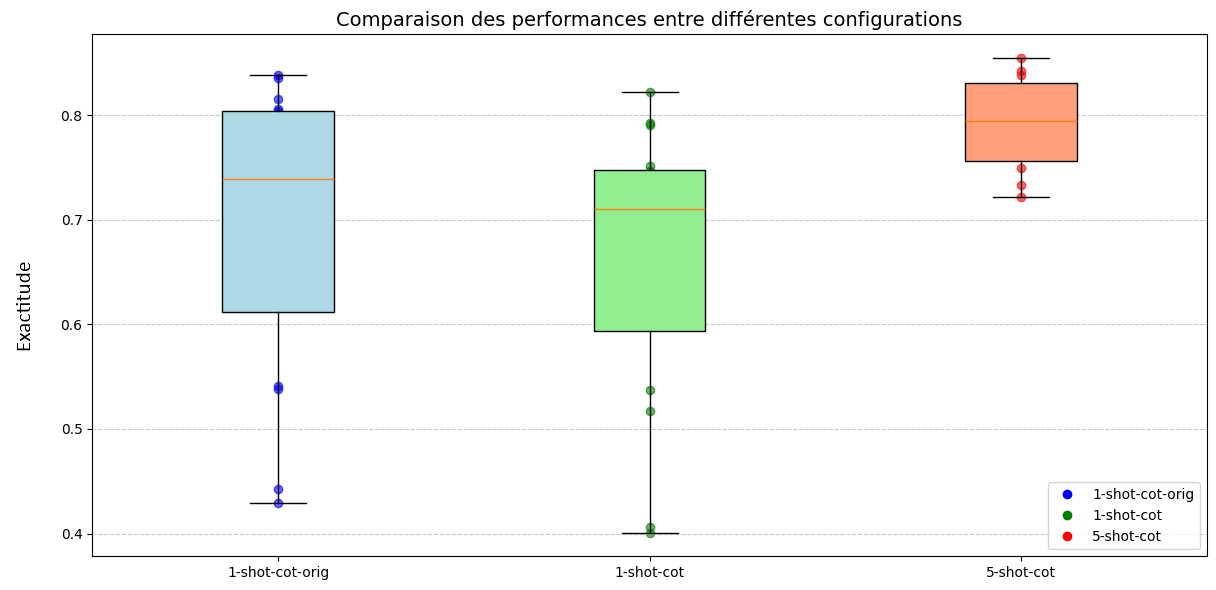}  
    \legend{Each point represents the performance of a model, and the boxes represent the quantiles (between 25~\% and 75~\%) and the median (orange line).}
    \caption{Comparison of accuracy for the use of human explanations (1-shot-cot-orig) versus the use of automatically generated explanations (1-shot-cot and 5-shot-cot)}
    \label{fig:man-gen_comparison}
\end{figure}

Finally, in Figure~\ref{fig:man-gen_comparison}, we observe the impact of using automatically generated explanations for methods that add reasoning (\textit{CoT}).

\paragraph{Effect of the presence or absence of automatically generated explanations} when comparing the \textit{1-shot-cot} and \textit{1-shot-cot-orig} methods, we notice a reduction in performance when using automatically generated annotations. Nevertheless, these automatic annotations allow for an overall improvement because they give us access to the \textit{5-shot-cot} method, which achieves the best performance.

\subsection{Detailed performance by CONSORT criterion}

\begin{figure}[!ht]
    \begin{subfigure}[b]{.5\textwidth}
        \centering
        \includegraphics[width=0.9\linewidth]{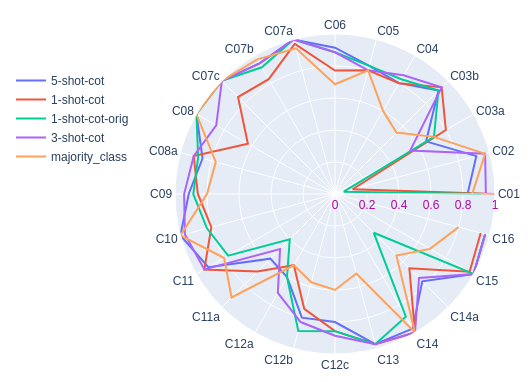}  
        \captionsetup{justification=centering,singlelinecheck=false}
        \caption{\textnormal{\textsc{CONSORT-QA-Covid}}}
        \label{fig:per_crit_covid}
    \end{subfigure}
    \begin{subfigure}[b]{.5\textwidth}
        \centering
        \includegraphics[width=0.9\linewidth]{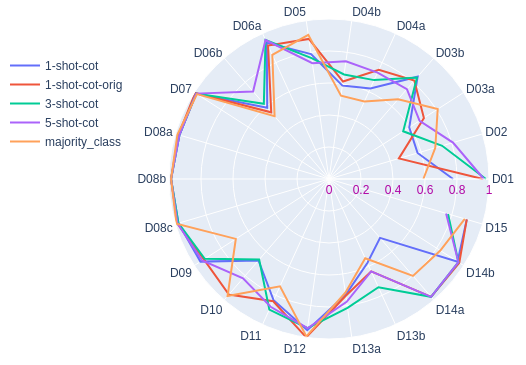}  
        \captionsetup{justification=centering,singlelinecheck=false}
        \caption{\textnormal{\textsc{CONSORT-QA-Dépression}}}
        \label{fig:per_crit_depression}
    \end{subfigure}
    \legend{The performance of the majority class on the test game is also displayed in orange.}
    \caption{Accuracy evaluated for each criterion on our corpora for the Mixtral-8x22B model with our most effective in-context learning strategies overall}
    \label{fig:per_criteria_accuracy}
\end{figure}

Figure~\ref{fig:per_criteria_accuracy} shows each criterion's performance to observe the \textsc{CONSORT-abstract} criteria for which more sophisticated in-context learning methods improve performance. We only study our best model, \textit{Mixtral-8x22B}, on our two corpora. We also display the majority class for each criterion (however, this is the majority class on the test set, to which our models do not have access; they only have a maximum of five examples in their context).

We can see that the model achieves very high performance mainly for criteria with unbalanced classes (such as D07, D08a, D08b, D08c, C07c, C08). Indeed, certain criteria are always verified in our corpora, so the models may be biased in relation to the examples provided, which may also be biased in favor of the majority class (but we are not analyzing this phenomenon here). We also note that adding examples is mostly beneficial, even though these examples are generated automatically.

\subsection{Correlation with criterion difficulty}

\begin{figure}[!ht]
    \centering
    \includegraphics[width=\linewidth]{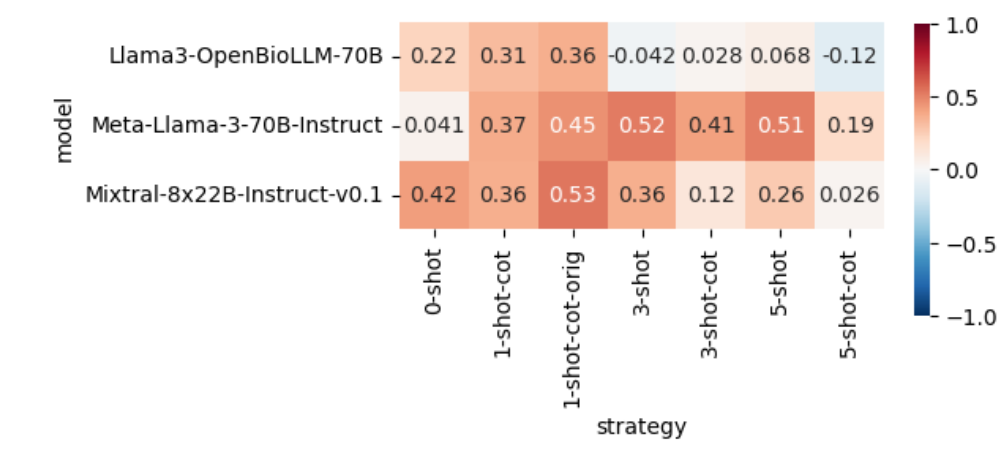}  
    \caption{Pearson correlation between model performance by criterion and the difficulty of these criteria}
    \label{fig:correlation}
\end{figure}

We calculate the correlation between the performance of the best model (\textit{Mixtral-8x22B} with the \textit{5-shot-cot} strategy) and the difficulty of the criteria, which we define by the inter-annotator agreement described in the study on depression (section~\ref{sec:corpus-description}), we therefore assume that if there is greater disagreement between human annotators, this means that the description of the criterion is imprecise and hence more challenging to evaluate. To do this, we measure the Pearson correlation \citep{benestyPearsonCorrelationCoefficient2009} between the performance (average accuracy) on each criterion and the kappa values presented in Table~\ref{tab:corpus-kappa}. We calculate these correlations for the different in-context learning methods with the three best models and present them in Figure~\ref{fig:correlation}.

We note that, on average, performances using manual explanations from the original \textsc{CONSORT-abstract} criteria in the prompt correlate more with kappa than those using automatically generated explanations. Manual annotations thus influence the model to generate behavior more correlated with a human annotator's behavior. Methods without explanations (\textit{0-shot}, \textit{3-shot}, and \textit{5-shot}) also have a higher correlation than their equivalents with explanations. Finally, the more automatically annotated examples there are, the weaker the correlation. And having seen previously that these methods perform better, the models can better generalize over more diverse examples and better understand the task (they therefore have less difficulty with the more difficult criteria).

\subsection{Qualitative analysis of the model's explanation for correct answers}\label{sec:correlation-explanations}

For the \textit{few-shot-cot} method, we want to verify whether the correct answers found by the model correlate with the explanation provided. We therefore sample 50 true positives and 50 true negatives from the generations of the \textit{Mixtral-8x22B} model \citep{jiangMixtralExperts2024} using the \textit{5-shot-cot} method. We then manually annotate each of these examples in a binary manner: either we judge the explanation to be correct (i.e., consistent with the context, the CONSORT question provided as input, and the answer given by the model), or we judge it to be incorrect (i.e., inconsistent with the context, the question, or the answer given). This gives us 99 out of 100 responses where the explanation provided by the model is consistent. Therefore, the explanations provided via the few-shot-cot method appear to be broadly consistent with the context and the response given by the model when the model responds correctly. However, we only perform this annotation with a single annotator (M. L.-K.), a PhD student who is not specialized in the biomedical field but who has acquired sufficient knowledge of randomized controlled clinical trials to verify the CONSORT criteria at the abstract level. We provide some examples of correct explanations in Figure~\ref{fig:correct-gen-example}.

\sethlcolor{gray!30}
\begin{figure*}[!ht]
    \centering
    \begin{tabular}{p{0.95\textwidth}}
        \hline
        \textbf{Question} : For the primary outcome, is there a result for each group and the estimated effect size and its precision (e.g., 95\% CI ) ? (if one of them is missing, answer no) \textbf{Explanation :}  \hl{The results section does not provide a clear report of the estimated effect size and its precision for the primary outcome. It only mentions that 'No significant group differences were observed on primary or secondary emotional outcomes' but does not provide the specific effect sizes and their precision.} \textbf{Answer :} \hl{No} \\\hline
        \textbf{Question} : Are the primary outcomes clearly described for this trial in methods ? \textbf{Explanation :}  \hl{The primary outcome is clearly described in the methods section as 'the incidence of either laboratory-confirmed Covid-19 or illness compatible with Covid-19 within 14 days'.} \textbf{Answer :} \hl{Yes} \\\hline
        \textbf{Question} : Do authors describe if the program deliverer (caregiver) were blinded ? (answer yes only if caregivers are blinded, do not care about participants or outcome assessors) \textbf{Explanation :}  \hl{The text does not mention whether the program deliverers (school clinicians) were blinded to the group assignment.} \textbf{Answer :} \hl{No} \\\hline
    \end{tabular}
    \legend{The \hl{highlighted text} corresponds to the parts generated by the model. Only questions and explanations are considered to limit the size of the examples. However, these explanation examples have been manually validated with their full context. Their justification has been annotated here as correct.}
    \caption{Correct examples generated by the \textit{Mixtral-8x22B} model using the 5-shot-cot method}
    \label{fig:correct-gen-example}
\end{figure*}

\subsection{Qualitative analysis of the model's explanation for incorrect responses}\label{sec:error-analysis}

We also analyze the parts the model generates when it makes a mistake. We sample 50 false positives and 50 false negatives for the Mixtral-8x22B model using the 5-shot-cot method. We manually annotate these errors with three qualitative criteria: 

\begin{itemize}
    \item Consistency of the explanation with the question: this criterion checks the extent to which the explanation provided by the model demonstrates an understanding of the question. We define three levels for annotation: consistent, partially consistent, and inconsistent. For the partially consistent annotation, this means that the model provides a logical explanation in relation to the question, but either does not consider part of the question or adds irrelevant elements.
    \item Consistency of the explanation with the context: we use the same annotation levels as for the previous criterion, but this time, we observe when the model cites part of the context to justify its response, or when it says that a criterion is missing, we check whether it is indeed missing from the abstract. If the model adds parts that do not exist (model hallucination), we immediately consider this an inconsistency (even if the non-existent part is unimportant for the response).
    \item consistency of the final answer with the explanation: here, we check whether the wording of the explanation is consistent with the final answer (criterion verified or not, i.e., generation of the \textit{token} "\textit{Yes}" or "\textit{No}"). The annotation "partially consistent" is specifically for cases where the model's explanation expresses both points of view (it finds the answer, but then nuances it and makes a mistake, for example). The inconsistent annotation refers to a correct explanation, but an incorrect final answer generation.
\end{itemize}

\begin{figure}[!ht]
    \centering
    \includegraphics[width=\linewidth]{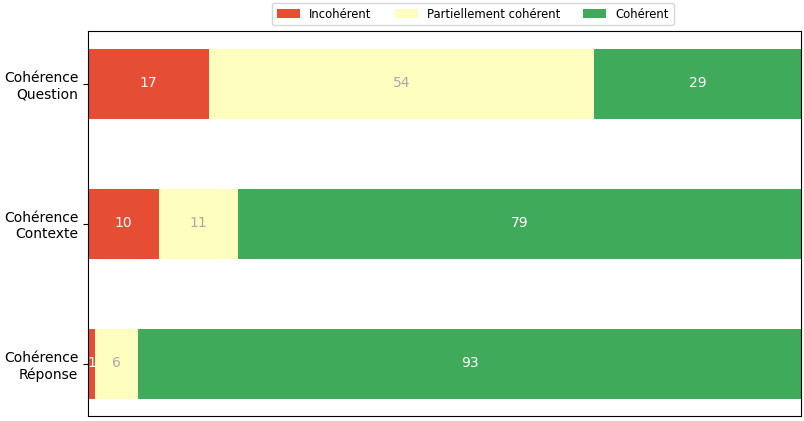}  
    \caption{Qualitative error analysis for explanations of the \textit{Mixtral-8x22B} model}
    \label{fig:error_analysis}
\end{figure}

We show the error analysis results in Figure~\ref{fig:error_analysis} and then provide examples of errors annotated in Figure~\ref{fig:error_examples}. Most of the model's errors stem from a misunderstanding of the question. Indeed, having chosen to use the definitions provided by the authors, we note that some are incomplete and may be imprecise, even for a human evaluator. We believe a more precise definition could improve performance on the criteria in question. For context errors, there are still sometimes hallucinations, where the model quotes parts that are not in the abstract. Limiting these errors by controlling the generation of the parts quoted by the model would also be possible.

\begin{figure*}[!tp]
    \centering
    \begin{tabular}{p{0.80\textwidth}|c|c|c}
        & Q & C & R \\
        \hline
        \textbf{Question :} Is there only a brief description of blinding (eg, single-blind, double-blind, triple-blind) ?
        \textbf{Explanation :} \hl{The authors report : 'Design : Parallel pilot randomized open label, double-masked clinical trial.' This means that the trial is not blinded at all, so there is a clear description of blinding (or lack thereof).}
        \textbf{Answer :} \hl{No} & \cellcolor{red!50}\xmark  & \cellcolor{green!50}\cmark & \cellcolor{yellow!50}$\pmb{\approx}$  \\
        \hline
        \textbf{Context :} Title : Structured Sensory Therapy (SITCAP-ART) for Traumatized Adjudicated Adolescents in Residential Treatment [...]
        \textbf{Question :} Is the study identified as randomized in the title ?
        \textbf{Explanation :} \hl{The title of this abstract ends with 'randomized controlled study'. It mentions the word 'randomized'. Hence, the study is correctly identified as randomized in the title}
        \textbf{Answer :} \hl{Yes} & \cellcolor{green!50}\cmark & \cellcolor{red!50}\xmark & \cellcolor{green!50}\cmark \\
        \hline
        \textbf{Question :} For the primary outcome(s), is the precision of the estimate (eg, 95\%CI) clearly stated ?
        \textbf{Explanation :} \hl{The primary outcome is not explicitly stated in the abstract. However, the abstract does provide the 95\% CI for the recovery rate ratio (1.87, 95\% CI, 0.72, 4.89) and the absolute risk reduction for invasive mechanical ventilation (32\%, 95\% CI, -0.07, 0.71).}
        \textbf{Answer :} \hl{Yes} & \cellcolor{green!50}\cmark & \cellcolor{yellow!50}$\pmb{\approx}$ & \cellcolor{red!50}\xmark \\
        \hline
    \end{tabular}
    \legend{The \hl{highlighted text} corresponds to the parts generated by the model. The columns Q, C, and R correspond to question, context, and response. The symbols mean: \xmark~=~inconsistent, \cmark~=~consistent, and $\pmb{\approx}$~=~partially consistent.}
    \caption{Incorrect examples generated by the Mixtral-8x22B model using the 5-shot-cot method, along with qualitative annotations}
    \label{fig:error_examples}
\end{figure*}

\subsection{Effect of filtering sentences according to the criterion}
We observe the influence of filtering abstract sentences only for the \textit{Llama-2-70b-chat} model, to limit the amount of computation required. The differences are minimal, as filtering is less effective on the various in-context learning strategies. Therefore, we can deduce that the model manages to retrieve information from an entire abstract despite the noise introduced by certain sentences irrelevant to the criterion being generated, or that the error introduced by the filtering model offsets the added benefit (but we do not analyze this phenomenon here). 

It is difficult to introduce this method into our comprehensive evaluation because the models used to filter sentences were not trained on precisely the same type of data (full texts, whereas we are evaluating summaries) or for the same criteria (CONSORT-2010 criteria, whereas we are using \textsc{CONSORT-abstract} criteria, although there is some overlap). Nevertheless, this technique can be interesting for few-shot strategies when the prompt size exceeds the model's maximum context size, as it reduces the number of sentences to be included in the prompt (given that abstracts take up most of the prompt size, especially when using multiple examples).

\section{Discussion}\label{sec:ch4:discussion}
\subsection{Towards automated evaluation of report quality}\label{sec:discussion-positive}

Our work is the first to evaluate the performance of large language models in assessing the quality of clinical trial abstract reports. It is important to ensure the quality of abstracts, which are usually the first thing that clinicians read. Sometimes, they may even limit themselves to reading only the abstract. 

An editorial \citep{nashwanStreamliningSystematicReviews2023} highlights the potential for large language models to improve the process of assessing the quality of clinical trials. However, they point out that we should not rely solely on machine learning methods, but that they could reduce the workload if combined with human expertise.

We also provide two corpora from different types of clinical trials evaluating the \textsc{CONSORT-abstract} criteria. We achieve good performance for the best models tested with our \textit{Chain-of-Thought} strategy, which also allow for a more transparent response from the model (thanks to the explanation provided). Certain specific criteria are even ideally evaluated by large language models.

The large language models allow us here, with the same model and without additional training, to meet the full range of criteria defined by the CONSORT standards. We highlight the importance of model size and the in-context learning method used to solve this task. However, our experiments have not yet conclusively established the usefulness of adapting these models to the biomedical field.

\subsection{Limitations}

\paragraph{Corpus size and variety} The corpora presented are small and cover only two different areas of clinical trials. Having larger corpora and a wider variety of medical specialties would allow us to improve our methods, refine the models, and obtain a more comprehensive evaluation. Furthermore, these criteria apply only to abstracts. For a more detailed and thorough assessment of an article, it may be helpful to evaluate the full text of the article. 

\paragraph{Use of large generative models} Although generative models are generally very effective for response tasks, they are not necessarily relevant for all the criteria evaluated here. Indeed, some pretty simple criteria can most likely be detected with smaller, and therefore less expensive, models. 

\paragraph{In-context learning} Large generative language models are also known to hallucinate. In our case, they may provide a false explanation or even quote passages from the abstract that do not exist for \textit{Chain-of-Thought} methods. It should be noted that it would be possible to control this last point automatically, at least for verbatim quotes, by constraining the model so that it does not generate sequences of tokens already present in its context.

\section{Conclusion}\label{sec:ch4:conclusion}

In this article, we report our initial experiences using large language models to automatically assess the quality of research article abstracts following a clinical trial. We targeted the \textsc{CONSORT-abstract} quality criteria, as the biomedical community widely acknowledges them.

To evaluate our models, we extracted two evaluation corpora produced by expert human evaluators, comprising 139 abstracts covering two clinical domains. 

We use one of the publicly available collections of large language models and compare the effects of model size and different in-context learning strategies. We achieve our best performance using the largest model (70 billion parameters), fine-tuned, and with the chain-of-thought strategy (\textsc{cot}). However, the value of models fine-tuned for the biomedical domain remains to be proven for our task. The same applies to the use of abstract sentence filtering upstream of generation.

Our models achieve nearly 85~\% accuracy for both corpora, demonstrating promising new avenues of research for evaluating the quality of clinical trial reports using this type of model. This is particularly noteworthy given that they require no additional training and can justify their responses, which adds some explainability to the reasoning behind the generation of the answer.

However, there is still room for improvement in these methods, particularly because some criteria remain more difficult to evaluate than others, and these models are still prone to hallucinations. There are many avenues for future work: adding more data to train these models, applying such methods to other standards (particularly standards that use full-text articles, or other widely used standards such as PRISMA for systematic reviews), or adding constraints to limit hallucinations (for example, controlling generation when the model quotes a passage from the abstract).

\bibliography{acl_latex}

\begin{thebibliography}{43}
\providecommand{\natexlab}[1]{#1}

\bibitem[{Altman et~al.(2001)Altman, Schulz, Moher, Egger, Davidoff, Elbourne, G{\o}tzsche, and Lang}]{altmanRevisedCONSORTStatement2001}
Douglas~G. Altman, Kenneth~F. Schulz, David Moher, Matthias Egger, Frank Davidoff, Diana Elbourne, Peter~C. G{\o}tzsche, and Thomas Lang. 2001.
\newblock \href {https://doi.org/10.7326/0003-4819-134-8-200104170-00012} {The {{Revised CONSORT Statement}} for {{Reporting Randomized Trials}}: {{Explanation}} and {{Elaboration}}}.
\newblock \emph{Annals of Internal Medicine}, 134(8):663--694.

\bibitem[{Ankit~Pal(2024)}]{OpenBioLLMs}
Malaikannan~Sankarasubbu Ankit~Pal. 2024.
\newblock Openbiollms: Advancing open-source large language models for healthcare and life sciences.
\newblock Hugging Face.

\bibitem[{Begg et~al.(1996)Begg, Cho, Eastwood, Horton, Moher, Olkin, Pitkin, Rennie, Schulz, Simel, and Stroup}]{beggImprovingQualityReporting1996}
Colin Begg, Mildred Cho, Susan Eastwood, Richard Horton, David Moher, Ingram Olkin, Roy Pitkin, Drummond Rennie, Kenneth~F. Schulz, David Simel, and Donna~F. Stroup. 1996.
\newblock \href {https://doi.org/10.1001/jama.1996.03540080059030} {Improving the {{Quality}} of {{Reporting}} of {{Randomized Controlled Trials}}: {{The CONSORT Statement}}}.
\newblock \emph{JAMA}, 276(8):637--639.

\bibitem[{Benesty et~al.(2009)Benesty, Chen, Huang, and Cohen}]{benestyPearsonCorrelationCoefficient2009}
Jacob Benesty, Jingdong Chen, Yiteng Huang, and Israel Cohen. 2009.
\newblock \href {https://doi.org/10.1007/978-3-642-00296-0_5} {Pearson {{Correlation Coefficient}}}.
\newblock In Israel Cohen, Yiteng Huang, Jingdong Chen, and Jacob Benesty, editors, \emph{Noise {{Reduction}} in {{Speech Processing}}}, pages 1--4. Springer, Berlin, Heidelberg.

\bibitem[{Bero et~al.(2021)Bero, Lawrence, Leslie, Chiu, McDonald, Page, Grundy, Parker, Boughton, Kirkham, and Featherstone}]{beroCrosssectionalStudyPreprints2021}
Lisa Bero, Rosa Lawrence, Louis Leslie, Kellia Chiu, Sally McDonald, Matthew~J. Page, Quinn Grundy, Lisa Parker, Stephanie Boughton, Jamie~J. Kirkham, and Robin Featherstone. 2021.
\newblock \href {https://doi.org/10.1136/bmjopen-2021-051821} {Cross-sectional study of preprints and final journal publications from {{COVID-19}} studies: Discrepancies in results reporting and spin in interpretation}.
\newblock \emph{BMJ open}, 11(7):e051821.

\bibitem[{Boutron et~al.(2014)Boutron, Altman, Hopewell, Vera-Badillo, Tannock, and Ravaud}]{IBoutron2014}
I~Boutron, D~G Altman, S~Hopewell, F~Vera-Badillo, I~Tannock, and P~Ravaud. 2014.
\newblock Impact of spin in the abstracts of articles reporting results of randomized controlled trials in the field of cancer: the spiin randomized controlled trial.
\newblock \emph{J Clin Oncol.}, 32(36):4120--4126.
\newblock Doi: 10.1200/JCO.2014.56.7503, PMID: 25403215.

\bibitem[{Brown et~al.(2020)Brown, Mann, Ryder, Subbiah, Kaplan, Dhariwal, Neelakantan, Shyam, Sastry, Askell, Agarwal, {Herbert-Voss}, Krueger, Henighan, Child, Ramesh, Ziegler, Wu, Winter, Hesse, Chen, Sigler, Litwin, Gray, Chess, Clark, Berner, McCandlish, Radford, Sutskever, and Amodei}]{brownLanguageModelsAre2020a}
Tom Brown, Benjamin Mann, Nick Ryder, Melanie Subbiah, Jared~D Kaplan, Prafulla Dhariwal, Arvind Neelakantan, Pranav Shyam, Girish Sastry, Amanda Askell, Sandhini Agarwal, Ariel {Herbert-Voss}, Gretchen Krueger, Tom Henighan, Rewon Child, Aditya Ramesh, Daniel Ziegler, Jeffrey Wu, Clemens Winter, and 12 others. 2020.
\newblock Language {{Models}} are {{Few-Shot Learners}}.
\newblock In \emph{Advances in {{Neural Information Processing Systems}}}, volume~33, pages 1877--1901. Curran Associates, Inc.

\bibitem[{Chen et~al.(2023)Chen, Hernández-Cano, Romanou, Bonnet, Matoba, Salvi, Pagliardini, Fan, Köpf, Mohtashami, Sallinen, Sakhaeirad, Swamy, Krawczuk, Bayazit, Marmet, Montariol, Hartley, Jaggi, and Bosselut}]{chen2023meditron70b}
Zeming Chen, Alejandro Hernández-Cano, Angelika Romanou, Antoine Bonnet, Kyle Matoba, Francesco Salvi, Matteo Pagliardini, Simin Fan, Andreas Köpf, Amirkeivan Mohtashami, Alexandre Sallinen, Alireza Sakhaeirad, Vinitra Swamy, Igor Krawczuk, Deniz Bayazit, Axel Marmet, Syrielle Montariol, Mary-Anne Hartley, Martin Jaggi, and Antoine Bosselut. 2023.
\newblock Meditron-70b: Scaling medical pretraining for large language models.
\newblock ArXiv.

\bibitem[{Devlin et~al.(2019)Devlin, Chang, Lee, and Toutanova}]{devlinBERTPretrainingDeep2019}
Jacob Devlin, Ming-Wei Chang, Kenton Lee, and Kristina Toutanova. 2019.
\newblock \href {https://doi.org/10.18653/v1/N19-1423} {{{BERT}}: {{Pre-training}} of {{Deep Bidirectional Transformers}} for {{Language Understanding}}}.
\newblock In \emph{Proceedings of the 2019 {{Conference}} of the {{North American Chapter}} of the {{Association}} for {{Computational Linguistics}}: {{Human Language Technologies}}, {{Volume}} 1 ({{Long}} and {{Short Papers}})}, pages 4171--4186, {Minneapolis, Minnesota}. {Association for Computational Linguistics}.

\bibitem[{Hopewell et~al.(2008)Hopewell, Clarke, Moher, Wager, Middleton, Altman, Schulz, and Group}]{hopewellCONSORTReportingRandomized2008}
Sally Hopewell, Mike Clarke, David Moher, Elizabeth Wager, Philippa Middleton, Douglas~G. Altman, Kenneth~F. Schulz, and {and} the~CONSORT Group. 2008.
\newblock \href {https://doi.org/10.1371/journal.pmed.0050020} {{{CONSORT}} for {{Reporting Randomized Controlled Trials}} in {{Journal}} and {{Conference Abstracts}}: {{Explanation}} and {{Elaboration}}}.
\newblock \emph{PLOS Medicine}, 5(1):e20.

\bibitem[{Jardim et~al.(2022)Jardim, Rose, Ames, Echavez, {Van de Velde}, and Muller}]{jardimAutomatingRiskBias2022}
Patricia Sofia~Jacobsen Jardim, Christopher~James Rose, Heather~Melanie Ames, Jose Francisco~Meneses Echavez, Stijn {Van de Velde}, and Ashley~Elizabeth Muller. 2022.
\newblock \href {https://doi.org/10.1186/s12874-022-01649-y} {Automating risk of bias assessment in systematic reviews: A real-time mixed methods comparison of human researchers to a machine learning system}.
\newblock \emph{BMC Medical Research Methodology}, 22(1):167.

\bibitem[{Jiang et~al.(2024{\natexlab{a}})Jiang, Sablayrolles, Roux, Mensch, Savary, Bamford, Chaplot, de~las Casas, Hanna, Bressand, Lengyel, Bour, Lample, Lavaud, Saulnier, Lachaux, Stock, Subramanian, Yang, Antoniak, Scao, Gervet, Lavril, Wang, Lacroix, and Sayed}]{jiangMixtralExperts2024}
Albert~Q. Jiang, Alexandre Sablayrolles, Antoine Roux, Arthur Mensch, Blanche Savary, Chris Bamford, Devendra~Singh Chaplot, Diego de~las Casas, Emma~Bou Hanna, Florian Bressand, Gianna Lengyel, Guillaume Bour, Guillaume Lample, L{\'e}lio~Renard Lavaud, Lucile Saulnier, Marie-Anne Lachaux, Pierre Stock, Sandeep Subramanian, Sophia Yang, and 7 others. 2024{\natexlab{a}}.
\newblock Mixtral of {{Experts}}.
\newblock arXiv.
\newblock \emph{arXiv preprint}.

\bibitem[{Jiang et~al.(2024{\natexlab{b}})Jiang, Lan, Menke, Vorland, and Kilicoglu}]{jiangCONSORTTMTextClassification2024}
Lan Jiang, Mengfei Lan, Joe~D. Menke, Colby~J. Vorland, and Halil Kilicoglu. 2024{\natexlab{b}}.
\newblock \href {https://doi.org/10.1101/2024.03.31.24305138} {{{CONSORT-TM}}: {{Text}} classification models for assessing the completeness of randomized controlled trial publications}.

\bibitem[{Jin et~al.(2021)Jin, Pan, Oufattole, Weng, Fang, and Szolovits}]{jinWhatDiseaseDoes2021a}
Di~Jin, Eileen Pan, Nassim Oufattole, Wei-Hung Weng, Hanyi Fang, and Peter Szolovits. 2021.
\newblock \href {https://doi.org/10.3390/app11146421} {What {{Disease Does This Patient Have}}? {{A Large-Scale Open Domain Question Answering Dataset}} from {{Medical Exams}}}.
\newblock \emph{Applied Sciences}, 11(14):6421.

\bibitem[{Jin and Szolovits(2018)}]{jinPICOElementDetection2018}
Di~Jin and Peter Szolovits. 2018.
\newblock \href {https://doi.org/10.18653/v1/W18-2308} {{{PICO Element Detection}} in {{Medical Text}} via {{Long Short-Term Memory Neural Networks}}}.
\newblock In \emph{Proceedings of the {{BioNLP}} 2018 Workshop}, pages 67--75, {Melbourne, Australia}. {Association for Computational Linguistics}.

\bibitem[{Kilicoglu et~al.(2021)Kilicoglu, Rosemblat, Hoang, Wadhwa, Peng, Mali{\v c}ki, Schneider, and {ter Riet}}]{kilicogluAssessingClinicalTrial2021}
Halil Kilicoglu, Graciela Rosemblat, Linh Hoang, Sahil Wadhwa, Zeshan Peng, Mario Mali{\v c}ki, Jodi Schneider, and Gerben {ter Riet}. 2021.
\newblock \href {https://doi.org/10.1016/j.jbi.2021.103717} {Toward assessing clinical trial publications for reporting transparency}.
\newblock \emph{Journal of Biomedical Informatics}, 116:103717.

\bibitem[{Kojima et~al.(2022)Kojima, Gu, Reid, Matsuo, and Iwasawa}]{kojimaLargeLanguageModels2022a}
Takeshi Kojima, Shixiang~(Shane) Gu, Machel Reid, Yutaka Matsuo, and Yusuke Iwasawa. 2022.
\newblock \href {https://proceedings.neurips.cc/paper_files/paper/2022/file/8bb0d291acd4acf06ef112099c16f326-Paper-Conference.pdf} {Large language models are zero-shot reasoners}.
\newblock In \emph{Advances in Neural Information Processing Systems}, volume~35, pages 22199--22213. {Curran Associates, Inc.}

\bibitem[{Koroleva(2020)}]{korolevaAssistedAuthoringAvoiding2020}
Anna Koroleva. 2020.
\newblock \emph{Assisted Authoring for Avoiding Inadequate Claims in Scientific Reporting}.
\newblock Ph.D. thesis, Universiteit von Amsterdam.

\bibitem[{Kwon et~al.(2023)Kwon, Li, Zhuang, Sheng, Zheng, Yu, Gonzalez, Zhang, and Stoica}]{kwon2023efficient}
Woosuk Kwon, Zhuohan Li, Siyuan Zhuang, Ying Sheng, Lianmin Zheng, Cody~Hao Yu, Joseph~E. Gonzalez, Hao Zhang, and Ion Stoica. 2023.
\newblock Efficient memory management for large language model serving with pagedattention.
\newblock In \emph{Proceedings of the ACM SIGOPS 29th Symposium on Operating Systems Principles}.

\bibitem[{Labrak et~al.(2024)Labrak, Bazoge, Morin, Gourraud, Rouvier, and Dufour}]{labrakBioMistralCollectionOpenSource2024a}
Yanis Labrak, Adrien Bazoge, Emmanuel Morin, Pierre-Antoine Gourraud, Mickael Rouvier, and Richard Dufour. 2024.
\newblock \href {https://doi.org/10.18653/v1/2024.findings-acl.348} {{{BioMistral}}: {{A Collection}} of {{Open-Source Pretrained Large Language Models}} for {{Medical Domains}}}.
\newblock In \emph{Findings of the {{Association}} for {{Computational Linguistics}}: {{ACL}} 2024}, pages 5848--5864, Bangkok, Thailand. Association for Computational Linguistics.

\bibitem[{Laï-king and Paroubek(2024)}]{laiking-paroubek-consort-qa}
Mathieu Laï-king and Patrick Paroubek. 2024.
\newblock \href {https://aclanthology.org/2022.tal-1.1/} {Évaluation de la qualité de rapport des essais cliniques avec des larges modèles de langue.}
\newblock In \emph{Traitement Automatique des Langues, Volume 65, Num{\'e}ro 2 :Traitement automatique de documents scientifiques}, pages 13--38, France. ATALA (Association pour le Traitement Automatique des Langues).

\bibitem[{Lee et~al.(2020)Lee, Yoon, Kim, Kim, Kim, So, and Kang}]{lee2020biobert}
Jinhyuk Lee, Wonjin Yoon, Sungdong Kim, Donghyeon Kim, Sunkyu Kim, Chan~Ho So, and Jaewoo Kang. 2020.
\newblock Biobert: a pre-trained biomedical language representation model for biomedical text mining.
\newblock \emph{Bioinformatics}, 36(4):1234--1240.

\bibitem[{Luo et~al.(2022)Luo, Sun, Xia, Qin, Zhang, Poon, and Liu}]{luoBioGPTGenerativePretrained2022}
Renqian Luo, Liai Sun, Yingce Xia, Tao Qin, Sheng Zhang, Hoifung Poon, and Tie-Yan Liu. 2022.
\newblock \href {https://doi.org/10.1093/bib/bbac409} {{BioGPT: generative pre-trained transformer for biomedical text generation and mining}}.
\newblock \emph{Briefings in Bioinformatics}, 23(6):bbac409.

\bibitem[{Marshall et~al.(2015)Marshall, Kuiper, and Wallace}]{marshallAutomatingRiskBias2015}
Iain~J. Marshall, Jo{\"e}l Kuiper, and Byron~C. Wallace. 2015.
\newblock \href {https://doi.org/10.1109/JBHI.2015.2431314} {Automating risk of bias assessment for clinical trials}.
\newblock \emph{IEEE journal of biomedical and health informatics}, 19(4):1406--1412.

\bibitem[{Marshall et~al.(2016)Marshall, Kuiper, and Wallace}]{marshallRobotReviewerEvaluationSystem2016}
Iain~J. Marshall, Jo{\"e}l Kuiper, and Byron~C. Wallace. 2016.
\newblock \href {https://doi.org/10.1093/jamia/ocv044} {{{RobotReviewer}}: Evaluation of a system for automatically assessing bias in clinical trials}.
\newblock \emph{Journal of the American Medical Informatics Association: JAMIA}, 23(1):193--201.

\bibitem[{Moher et~al.(2010)Moher, Hopewell, Schulz, Montori, Gøtzsche, Devereaux, Elbourne, Egger, and Altman}]{moherCONSORT2010Explanation2010}
David Moher, Sally Hopewell, Kenneth~F. Schulz, Victor Montori, Peter~C. Gøtzsche, P.~J. Devereaux, Diana Elbourne, Matthias Egger, and Douglas~G. Altman. 2010.
\newblock \href {https://doi.org/10.1136/bmj.c869} {{{CONSORT}} 2010 {{Explanation}} and {{Elaboration}}: Updated guidelines for reporting parallel group randomised trials}.
\newblock \emph{BMJ}, 340:c869.

\bibitem[{Mutinda et~al.(2022)Mutinda, Liew, Yada, Wakamiya, and Aramaki}]{mutindaAutomaticDataExtraction2022}
Faith~Wavinya Mutinda, Kongmeng Liew, Shuntaro Yada, Shoko Wakamiya, and Eiji Aramaki. 2022.
\newblock \href {https://doi.org/10.1186/s12911-022-01897-4} {Automatic data extraction to support meta-analysis statistical analysis: A case study on breast cancer}.
\newblock \emph{BMC Medical Informatics and Decision Making}, 22:158.

\bibitem[{Nashwan et~al.(2023)Nashwan, Jaradat, Nashwan, and Jaradat}]{nashwanStreamliningSystematicReviews2023}
Abdulqadir~J. Nashwan, Jaber~H. Jaradat, Abdulqadir~J. Nashwan, and Jaber~H. Jaradat. 2023.
\newblock \href {https://doi.org/10.7759/cureus.43023} {Streamlining {{Systematic Reviews}}: {{Harnessing Large Language Models}} for {{Quality Assessment}} and {{Risk-of-Bias Evaluation}}}.
\newblock \emph{Cureus}, 15(8).

\bibitem[{Neumann et~al.(2019)Neumann, King, Beltagy, and Ammar}]{neumann-etal-2019-scispacy}
Mark Neumann, Daniel King, Iz~Beltagy, and Waleed Ammar. 2019.
\newblock \href {https://doi.org/10.18653/v1/W19-5034} {{S}cispa{C}y: {F}ast and {R}obust {M}odels for {B}iomedical {N}atural {L}anguage {P}rocessing}.
\newblock In \emph{Proceedings of the 18th BioNLP Workshop and Shared Task}, pages 319--327, Florence, Italy. Association for Computational Linguistics.

\bibitem[{Singhal et~al.(2023)Singhal, Azizi, Tu, Mahdavi, Wei, Chung, Scales, Tanwani, Cole-Lewis, Pfohl, Payne, Seneviratne, Gamble, Kelly, Babiker, Schärli, Chowdhery, Mansfield, Demner-Fushman, Agüera~y Arcas, Webster, Corrado, Matias, Chou, Gottweis, Tomasev, Liu, Rajkomar, Barral, Semturs, Karthikesalingam, and Natarajan}]{singhalLargeLanguageModels2023}
Karan Singhal, Shekoofeh Azizi, Tao Tu, S.~Sara Mahdavi, Jason Wei, Hyung~Won Chung, Nathan Scales, Ajay Tanwani, Heather Cole-Lewis, Stephen Pfohl, Perry Payne, Martin Seneviratne, Paul Gamble, Chris Kelly, Abubakr Babiker, Nathanael Schärli, Aakanksha Chowdhery, Philip Mansfield, Dina Demner-Fushman, and 13 others. 2023.
\newblock \href {https://doi.org/10.1038/s41586-023-06291-2} {Large language models encode clinical knowledge}.
\newblock \emph{Nature}, 620(7972):172--180.

\bibitem[{Singhal et~al.(2025)Singhal, Tu, Gottweis, Sayres, Wulczyn, Amin, Hou, Clark, Pfohl, {Cole-Lewis}, Neal, Rashid, Schaekermann, Wang, Dash, Chen, Shah, Lachgar, Mansfield, Prakash, Green, Dominowska, {Ag{\"u}era y Arcas}, Toma{\v s}ev, Liu, Wong, Semturs, Mahdavi, Barral, Webster, Corrado, Matias, Azizi, Karthikesalingam, and Natarajan}]{singhalExpertLevelMedicalQuestion2025}
Karan Singhal, Tao Tu, Juraj Gottweis, Rory Sayres, Ellery Wulczyn, Mohamed Amin, Le~Hou, Kevin Clark, Stephen~R. Pfohl, Heather {Cole-Lewis}, Darlene Neal, Qazi~Mamunur Rashid, Mike Schaekermann, Amy Wang, Dev Dash, Jonathan~H. Chen, Nigam~H. Shah, Sami Lachgar, Philip~Andrew Mansfield, and 16 others. 2025.
\newblock \href {https://doi.org/10.1038/s41591-024-03423-7} {Toward expert-level medical question answering with large language models}.
\newblock \emph{Nature Medicine}, 31(3):943--950.

\bibitem[{Sterne et~al.(2019)Sterne, Savovi{\'c}, Page, Elbers, Blencowe, Boutron, Cates, Cheng, Corbett, Eldridge, Emberson, Hern{\'a}n, Hopewell, Hr{\'o}bjartsson, Junqueira, J{\"u}ni, Kirkham, Lasserson, Li, McAleenan, Reeves, Shepperd, Shrier, Stewart, Tilling, White, Whiting, and Higgins}]{sterneRoBRevisedTool2019}
Jonathan A.~C. Sterne, Jelena Savovi{\'c}, Matthew~J. Page, Roy~G. Elbers, Natalie~S. Blencowe, Isabelle Boutron, Christopher~J. Cates, Hung-Yuan Cheng, Mark~S. Corbett, Sandra~M. Eldridge, Jonathan~R. Emberson, Miguel~A. Hern{\'a}n, Sally Hopewell, Asbj{\o}rn Hr{\'o}bjartsson, Daniela~R. Junqueira, Peter J{\"u}ni, Jamie~J. Kirkham, Toby Lasserson, Tianjing Li, and 9 others. 2019.
\newblock \href {https://doi.org/10.1136/bmj.l4898} {{{RoB}} 2: A revised tool for assessing risk of bias in randomised trials}.
\newblock \emph{BMJ}, 366:l4898.

\bibitem[{Touvron et~al.(2023{\natexlab{a}})Touvron, Lavril, Izacard, Martinet, Lachaux, Lacroix, Rozi{\`e}re, Goyal, Hambro, Azhar, Rodriguez, Joulin, Grave, and Lample}]{touvronLLaMAOpenEfficient2023}
Hugo Touvron, Thibaut Lavril, Gautier Izacard, Xavier Martinet, Marie-Anne Lachaux, Timoth{\'e}e Lacroix, Baptiste Rozi{\`e}re, Naman Goyal, Eric Hambro, Faisal Azhar, Aurelien Rodriguez, Armand Joulin, Edouard Grave, and Guillaume Lample. 2023{\natexlab{a}}.
\newblock \href {https://doi.org/10.48550/arXiv.2302.13971} {{{LLaMA}}: {{Open}} and {{Efficient Foundation Language Models}}}.
\newblock arXiv.
\newblock \emph{arXiv preprint}.

\bibitem[{Touvron et~al.(2023{\natexlab{b}})Touvron, Martin, Stone, Albert, Almahairi, Babaei, Bashlykov, Batra, Bhargava, Bhosale, Bikel, Blecher, Ferrer, Chen, Cucurull, Esiobu, Fernandes, Fu, Fu, Fuller, Gao, Goswami, Goyal, Hartshorn, Hosseini, Hou, Inan, Kardas, Kerkez, Khabsa, Kloumann, Korenev, Koura, Lachaux, Lavril, Lee, Liskovich, Lu, Mao, Martinet, Mihaylov, Mishra, Molybog, Nie, Poulton, Reizenstein, Rungta, Saladi, Schelten, Silva, Smith, Subramanian, Tan, Tang, Taylor, Williams, Kuan, Xu, Yan, Zarov, Zhang, Fan, Kambadur, Narang, Rodriguez, Stojnic, Edunov, and Scialom}]{touvronLlamaOpenFoundation2023}
Hugo Touvron, Louis Martin, Kevin Stone, Peter Albert, Amjad Almahairi, Yasmine Babaei, Nikolay Bashlykov, Soumya Batra, Prajjwal Bhargava, Shruti Bhosale, Dan Bikel, Lukas Blecher, Cristian~Canton Ferrer, Moya Chen, Guillem Cucurull, David Esiobu, Jude Fernandes, Jeremy Fu, Wenyin Fu, and 49 others. 2023{\natexlab{b}}.
\newblock \href {https://doi.org/10.48550/arXiv.2307.09288} {Llama 2: {{Open Foundation}} and {{Fine-Tuned Chat Models}}}.
\newblock arXiv.
\newblock \emph{arXiv preprint}.

\bibitem[{Turner et~al.(2012)Turner, Shamseer, Altman, Schulz, and Moher}]{turnerDoesUseCONSORT2012}
Lucy Turner, Larissa Shamseer, Douglas~G. Altman, Kenneth~F. Schulz, and David Moher. 2012.
\newblock \href {https://doi.org/10.1186/2046-4053-1-60} {Does use of the {{CONSORT Statement}} impact the completeness of reporting of randomised controlled trials published in medical journals? {{A Cochrane}} review}.
\newblock \emph{Systematic Reviews}, 1:60.

\bibitem[{Vaswani et~al.(2017)Vaswani, Shazeer, Parmar, Uszkoreit, Jones, Gomez, Kaiser, and Polosukhin}]{vaswaniAttentionAllYou2017a}
Ashish Vaswani, Noam Shazeer, Niki Parmar, Jakob Uszkoreit, Llion Jones, Aidan~N Gomez, Lukasz Kaiser, and Illia Polosukhin. 2017.
\newblock \href {https://papers.nips.cc/paper_files/paper/2017/hash/3f5ee243547dee91fbd053c1c4a845aa-Abstract.html} {Attention is {{All}} you {{Need}}}.
\newblock In \emph{Advances in {{Neural Information Processing Systems}}}, volume~30. {Curran Associates, Inc.}

\bibitem[{Wang et~al.(2021)Wang, Chen, Wang, Hua, Li, Li, Zhang, Fan, Li, and Clarke}]{wangAbstractsReportsRandomized2021}
Dongguang Wang, Lingmin Chen, Lian Wang, Fang Hua, Juan Li, Yuxi Li, Yonggang Zhang, Hong Fan, Weimin Li, and Mike Clarke. 2021.
\newblock \href {https://doi.org/10.1016/j.jclinepi.2021.06.027} {Abstracts for reports of randomized trials of {{COVID-19}} interventions had low quality and high spin}.
\newblock \emph{Journal of Clinical Epidemiology}, 139:107--120.

\bibitem[{Wang et~al.(2020)Wang, Schilsky, Page, Califf, Cheung, Wang, and Pang}]{wangDevelopmentValidationNatural2020}
Fan Wang, Richard~L. Schilsky, David Page, Robert~M. Califf, Kei Cheung, Xiaofei Wang, and Herbert Pang. 2020.
\newblock \href {https://doi.org/10.1001/jamanetworkopen.2020.14661} {Development and {{Validation}} of a {{Natural Language Processing Tool}} to {{Generate}} the {{CONSORT Reporting Checklist}} for {{Randomized Clinical Trials}}}.
\newblock \emph{JAMA Network Open}, 3(10):e2014661.

\bibitem[{Wang et~al.(2022)Wang, Liao, Lapata, and Macleod}]{wangPICOEntityExtraction2022}
Qianying Wang, Jing Liao, Mirella Lapata, and Malcolm Macleod. 2022.
\newblock \href {https://doi.org/10.1186/s13643-022-02074-4} {{{PICO}} entity extraction for preclinical animal literature}.
\newblock \emph{Systematic Reviews}, 11(1):209.

\bibitem[{Warrier and Jayanthi(2022)}]{warrierCompletenessReportingOutcome2022}
Kiran Warrier and C.~R. Jayanthi. 2022.
\newblock \href {https://doi.org/10.4103/picr.PICR_64_20} {Completeness of reporting and outcome switching in trials published in {{Indian}} journals from 2017 to 2019: {{A}} cross-sectional study}.
\newblock \emph{Perspectives in Clinical Research}, 13(2):77--81.

\bibitem[{Wei et~al.(2022)Wei, Wang, Schuurmans, Bosma, Ichter, Xia, Chi, Le, and Zhou}]{weiChainofThoughtPromptingElicits2022}
Jason Wei, Xuezhi Wang, Dale Schuurmans, Maarten Bosma, Brian Ichter, Fei Xia, Ed~Chi, Quoc~V. Le, and Denny Zhou. 2022.
\newblock \href {https://proceedings.neurips.cc/paper_files/paper/2022/hash/9d5609613524ecf4f15af0f7b31abca4-Abstract-Conference.html} {Chain-of-{{Thought Prompting Elicits Reasoning}} in {{Large Language Models}}}.
\newblock \emph{Advances in Neural Information Processing Systems}, 35:24824--24837.

\bibitem[{Wiehn et~al.(2022)Wiehn, Nonte, and Prugger}]{wiehnReportingQualityAbstracts2022}
Jascha Wiehn, Johanna Nonte, and Christof Prugger. 2022.
\newblock \href {https://doi.org/10.1136/bmjopen-2022-061873} {Reporting quality for abstracts of randomised trials on child and adolescent depression prevention: A meta-epidemiological study on adherence to {{CONSORT}} for abstracts}.
\newblock \emph{BMJ Open}, 12(8):e061873.

\bibitem[{Yang et~al.(2023)Yang, Wang, Lu, Liu, Le, Zhou, and Chen}]{yangLargeLanguageModels2023}
Chengrun Yang, Xuezhi Wang, Yifeng Lu, Hanxiao Liu, Quoc~V. Le, Denny Zhou, and Xinyun Chen. 2023.
\newblock \href {https://arxiv.org/abs/2309.03409} {Large {{Language Models}} as {{Optimizers}}}.
\newblock \emph{arXiv:2309.03409}.

\end{thebibliography}

\end{document}